\documentclass[10pt,twocolumn,letterpaper]{article}

\usepackage{ijcb}
\usepackage{times}
\usepackage{epsfig}
\usepackage{graphicx}
\usepackage{amsmath}
\usepackage{amssymb}

\usepackage{cite}
\usepackage{amsmath,amssymb,amsfonts}
\usepackage{algorithm, algorithmic}
\usepackage{booktabs}
\usepackage{multirow}
\usepackage{graphicx}
\usepackage{url} 
\usepackage{textcomp}
\usepackage{xcolor}
\usepackage{mathtools}

\DeclareUnicodeCharacter{2212}{-} 

\ijcbfinalcopy 

\ifijcbfinal\fi
\begin{document}

\title{Specular- and Diffuse-reflection-based\\Face Spoofing Detection for Mobile Devices}

\author{Akinori F. Ebihara\textsuperscript{1} \hspace{2cm} Kazuyuki Sakurai\textsuperscript{1} \hspace{2cm} Hitoshi Imaoka\textsuperscript{2}\\
\textsuperscript{1}NEC Biometrics Research Laboratories \hspace{2cm} \textsuperscript{2}NEC Corporation\\
{\tt\small aebihara@nec.com}
}
\maketitle

\thispagestyle{empty}

\begin{abstract}
In light of the rising demand for biometric-authentication systems, preventing face spoofing attacks is a critical issue for the safe deployment of face recognition systems. Here, we propose an efficient face presentation attack detection (PAD) algorithm that requires minimal hardware and only a small database, making it suitable for resource-constrained devices such as mobile phones. Utilizing one monocular visible light camera, the proposed algorithm takes two facial photos, one taken with a flash, the other without a flash. The proposed $SpecDiff$ descriptor is constructed by leveraging two types of reflection: (i) specular reflections from the iris region that have a specific intensity distribution depending on liveness, and (ii) diffuse reflections from the entire face region that represents the 3D structure of a subject's face. Classifiers trained with $SpecDiff$ descriptor outperforms other flash-based PAD algorithms on both an in-house database and on publicly available NUAA, Replay-Attack, and SiW databases. Moreover, the proposed algorithm achieves statistically significantly better accuracy to that of an end-to-end, deep neural network classifier, while being approximately six-times faster execution speed. The code is publicly available at \url{https://github.com/Akinori-F-Ebihara/SpecDiff-spoofing-detector}.
\end{abstract}

\section{Introduction}
\begin{figure}[t]
\centerline{\includegraphics[keepaspectratio]{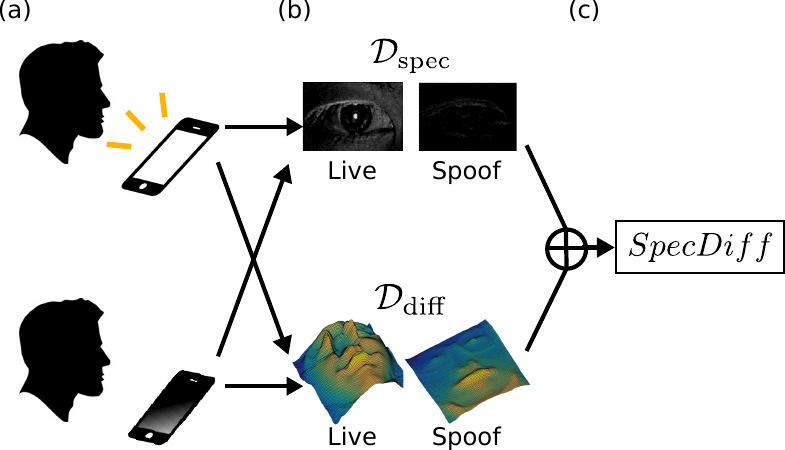}}
\caption{Construction of the proposed $SpecDiff$ descriptor. (a) Two photos, one taken with a flash (top), the other without a flash (bottom), are shot within $\approx200$ milliseconds. (b) Face and eye regions are extracted from the two photos and resized, and the \textit{Speculum Descriptor} ($\mathcal{D}_{\mathrm{spec}} \in \mathbb{R}^{3200}$ for each eye) and the \textit{Diffusion Descriptor} ($\mathcal{D}_{\mathrm{diff}} \in \mathbb{R}^{10000}$) are calculated from these regions. Two example results obtained with live (left) and spoof (right) faces are shown. Note that the actual region used to calculate $D_{Spec}$ is $40\times40$ pixels square region inside the iris centered on the pupil. (c) The two descriptors are vectorized and concatenated to build the $SpecDiff$ descriptor, which is then classified by standard classifiers such as a support vector machine (SVM, \cite{SVMoriginal}) or a neural network.}
\label{fig:Conceptual}
\end{figure}
A biometric authentication system has an advantage over a traditional password-based authentication system: it uses intrinsic features such as a face or fingerprint, so the user does not have to remember anything to be authenticated. Among the various biometric authentication systems, face-recognition-based ones take advantage of the huge variety of facial features across individuals, and thus have the potential to offer convenience and high security. Face authentication, however, has a major drawback common to other forms of biometric authentication: a nonzero probability of false rejection and false acceptance. While false rejection is less problematic, because a genuine user can usually make a second attempt to be authorized, false acceptance entails a higher security risk. When a false acceptance occurs, the system may actually be under an attack by a malicious imposter attempting to break into it. Acquiring facial images via social networks is now easier than ever, allowing attackers to execute a variety of attacks using printed photos or recorded video. The demand for technologies for detecting face presentation attack detection (PAD) is thus rising in an effort to ensure the security of sites deploying face recognition systems. Face recognition systems are being used at, for example, airports and office entrances and as login systems of edge devices. Each site has its own hardware availability; i.e., it may have access to a server that can perform computationally expensive calculations, or it may be equipped with infrared imaging devices. On the other hand, it may only have access to a low-performance CPU. It is thus natural that the suitable face PAD algorithm will differ according to the hardware availability. The advent of deep-learning technologies has allowed high-precision image processing that competes with human abilities at the expense of high computational cost. On the other hand, there is still a need for an efficient PAD algorithm that works with minimal computational resources. In this study, we focus on this case: PAD on a mobile phone equipped only with a CPU, without access to external servers.\\
\indent In line the goal of developing PAD technology independent of hardware requirements, we decided to use one visible-light camera mounted on the front of the mobile device, and have devised an efficient, novel flash reflection-based face PAD algorithm.
\begin{figure}[tb]
\centerline{\includegraphics[keepaspectratio]{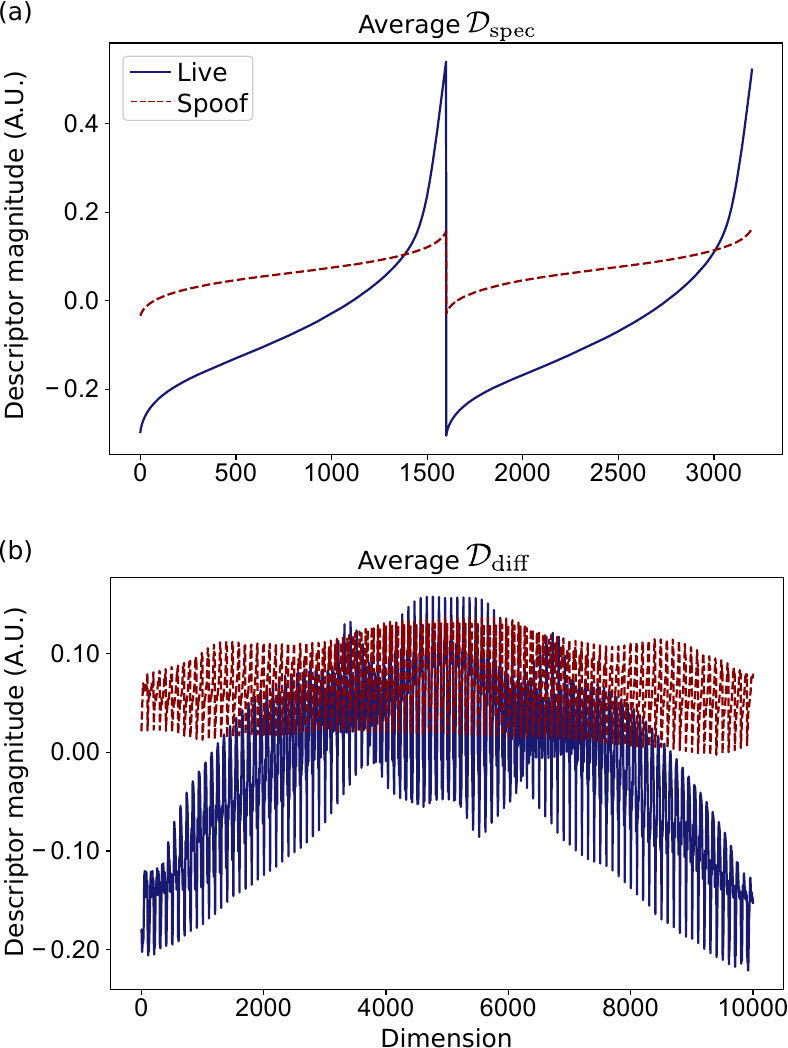}}
\caption{The proposed $\mathcal{D}_{\mathrm{spec}}$ and $\mathcal{D}_{\mathrm{diff}}$, calculated and averaged across the live face class ($N=1176$, blue) and the spoof face class ($N=1660$, red). The ordinates have arbitrary units (A.U.). (a) Vectorized \textit{Speculum Descriptor}, $\mathcal{D}_{\mathrm{spec}}\in\mathbb{R}^{3200}$, calculated from iris regions. (b) Vectorized \textit{Diffusion Descriptor}, $\mathcal{D}_{\mathrm{diff}}\in\mathbb{R}^{10000}$, calculated from face regions.}
\label{fig:descriptor}
\end{figure}
The algorithm leverages specular and diffuse reflections from the iris regions and the facial surface, respectively. An all-white, bright screen is used as a flash simulator, and two facial photos, taken with and without a flash, are used to calculate the \textit{Speculum Descriptor}, $\mathcal{D}_{\mathrm{spec}} \in \mathbb{R}^{3200}$ and the \textit{Diffusion Descriptor}, $\mathcal{D}_{\mathrm{diff}} \in \mathbb{R}^{10000}$. Both descriptors are based on the difference between the two facial photos and are normalized by the luminance intensities such that the descriptor magnitude is bounded in the range $[-1, 1]$, thereby facilitating the training of classifiers and improvement of their classification accuracy (Fig.\ref{fig:descriptor}).\\
\indent Testing on a small in-house database containing $\approx 1K$ image pairs per binary class (live or spoof face), the proposed descriptor classified with a support vector machine (SVM) achieved the highest classification performance among other flash-based face PAD algorithms. Generalizability across different domains is verified by cross-database evaluation on NUAA, Replay-Attack, and SiW databases; all classifiers are trained on the in-house database and tested on the three public databases. The results confirmed that the proposed algorithm not only outperforms other flash-based algorithms on the public databases but also achieves statistically significantly better classification performance than that of a computationally expensive, end-to-end deep neural network that is six-times slower.\\
\indent The proposed algorithm enables efficient, user-friendly, and accurate PAD. Its contributions are summarized below:

\begin{enumerate}
\item Minimal hardware requirements: a single visible-light camera and a flash light-emitting device.
\item Minimal computational requirements: implementable on mobile devices.
\item Minimal database requirements: trainable with merely $\approx1K$ image pairs for both live and spoof face classes.
\item Minimal data label requirements: no auxiliary supervision such as depth or segmentation is needed.
\item High detection accuracy, better than an end-to-end, deep neural network model, but with six-times faster execution.
\end{enumerate}
\section{Related Work}
The current PAD technologies aimed against spoofing attacks are summarized below. Face spoofing attacks can be subdivided into two major categories: 2D attacks and 3D attacks. The former includes print-attacks and video-replay attacks, while the latter includes 3D spoofing mask attacks. Several publicly available databases simulate these attacks. To name a few, the NUAA \cite{NUAA} and Print-Attack \cite{PrintAttack} databases simulate photo attacks. The Replay-Attack \cite{ReplayAttack}, CASIA Face Anti-Spoofing \cite{CASIA}, The MSU Mobile Face Spoofing Database \cite{MSU_MFSD}, and Spoofing in the Wild (SiW, \cite{SiW}) databases contain replay attacks in addition to photo attacks. The 3D Mask Attack Database\cite{3DMAD} and HKBU-Mask Attack with Real World Variations \cite{HKBUMARs} simulate 3D mask attacks. Example countermeasures to each attack type are summarized below.
\subsection{Countermeasures to 2D attacks}
Because of the reflectance of printed media and the use of photo compression, printed photos have surface textures or patterns that differ from those of a live human face, and these textures can be used to detect print attacks. Replay attacks are conducted by playing video on displays such as monitors or screens, which also have surface properties different from those of a live face. Here, local binary pattern (LBP,\cite{ReplayAttack, LBP1, LBP2}), Gaussian filtering\cite{Gauss1, Gauss2}, and their variants can be used to detect 2D attacks.\\
\indent Infrared imaging can be used to counter replay attacks, because the display emits light only at visible wavelengths (i.e., a face does not appear in an infrared picture taken of a display whereas it appears in an image of an actual person \cite{Song2018}). Another replay-attack-specific surface property is moir\'{e} pattern \cite{moire}.\\
\indent A prominent feature of these 2D attacks is the flat, 2D structure of the spoofing media. Here, stereo vision\cite{Stereo}, depth measurement from defocusing\cite{Focus}, and flash-based 3D measurements \cite{FlashLivenessChan,FlashLivenessTang,MDM,AuroraGuard} are effective countermeasures that detect flatness as a surrogate of 2D spoofing attacks. In this paper, we focus on using flash-based PAD to counter 2D attacks.\\
\indent Some algorithms, including ours, construct descriptors from pictures taken with or without a flash. The following four are mono-colored-flash-based algorithms: (i) LBP\textunderscore FI (LBP on the flash image), in which the LBP of a picture taken with a flash is used as a descriptor\cite{FlashLivenessChan}; (ii) SD\textunderscore FIC (the standard deviation of face intensity change), in which the standard deviation of the difference between photos of the same subject taken with and without a flash is used as a descriptor \cite{FlashLivenessChan}; (iii) Face flashing, in which the descriptor is made from the relative reflectance between two different pixels in one photo taken with a flash, i.e., the reflectance of each facial pixel divided by that of a reflectance pixel (hereafter abbreviated as RelativeRef \cite{FlashLivenessTang}); (iv) implicit 3D features, where pixel-wise differences in pictures taken with and without a flash are calculated and divided by the pixel intensity of the picture without the flash on a pixel-by-pixel basis \cite{MDM}. We compare these algorithms with ours in Results section.
\subsection{Countermeasures to 3D mask attacks}
The recent 3D reconstruction and printing technologies have given malicious users the ability to produce realistic spoofing masks\cite{MaskReview}. One example countermeasure against such a 3D attack is multispectral imaging. Steiner et al.\cite{SWIR} have reported the effectiveness of short-wave infrared (SWIR) imaging for detecting masks. Another approach is remote photoplethysmography (rPPG), which calculates pulse rhythms from periodic changes in face color \cite{rPPG}.\\ 
\indent In this paper, however, we do not consider 3D attacks because they are less likely due to the high cost of producing 3D masks. Our work focuses on preventing photo attacks and replay attacks.
\subsection{End-to-end deep neural networks}
The advent of deep learning has allowed researchers to construct an end-to-end classifier without having to design an explicit descriptor. Research on face PAD is no exception; that is, deep neural network-based countermeasures have been found for not only photo attacks but also replay and 3D mask attacks\cite{CNN1, CNN2, CNN3}. The Experiment section compares our algorithm's performance with that of a deep neural network-based, end-to-end classifier.
\section{Proposed algorithm}
We propose a PAD algorithm that uses both specular and diffuse reflection of flash light. The iris regions and the facial surface are used to compute the \textit{Speculum Descriptor}, $\mathcal{D}_{\mathrm{spec}}$, and the \textit{Diffusion Descriptor}, $\mathcal{D}_{\mathrm{diff}}$, respectively. The two descriptors are vectorized and concatenated to build the $SpecDiff$ descriptor, which can be classified as a live or spoof face by using a standard classifier such as SVM or a neural network. \\
\indent The procedure is as follows. During and after the flash illumination, two RGB, three-channel photos are taken: $P^{(\mathrm f)}$ with a flash, and $P^{(\mathrm b)}$ under background light. Since not all front cameras of smartphones are equipped with a flash LED, all display pixels are simultaneously excited with the highest luminance intensity to simulate light from a flash. The flash duration is as short as 200 milliseconds such that it does not harm the user experience. \\
\indent Before $\mathcal{D}_{\mathrm{spec}}$ and $\mathcal{D}_{\mathrm{diff}}$ are calculated, the following common preprocessing functions are applied to both $P^{(\mathrm f)}$ and $P^{(\mathrm b)}$:
\begin{itemize}
\item Function $Rotation$: make the facial image upright.
\item Function $GrayScale$: transfer RGB images into grayscale images.
\item Function $FaceDetection$: detect face location ($Face_{\mathrm{cen}}$) from each image.
\item Function $FeatExtraction$: extract locations of facial feature points ($Face_{\mathrm{loc}}$) from each face.
\item Function $FaceCrop$: crop the region of interest.
\item Function $GaussFilter$: apply Gaussian filter to increase position invariance.
\item Function $Resize$: resize the region of interest.
\end{itemize}
For face detection and facial feature extraction, the LBP-AdaBoost algorithm\cite{ViolaJones} and the supervised descent method\cite{SDM} are used in combination. Note that because Gaussian filtration is applied after cropping the region of interest, it does not take significant calculation time compared with the other processings. For the details of cropping, filtering, and resizing, see sections \ref{subsec:Dspec} and \ref{subsec:Ddiff}. \\
\indent The positions of the faces detected in the two photos may potentially be different. However, because the flash duration is as short as 200 ms, the positional difference does not cause a major problem with face alignment.
\subsection{$\mathcal{D}_{\mathrm{spec}}$: specular-reflection-based descriptor}\label{subsec:Dspec}
\begin{figure}[tb]
\centerline{\includegraphics[keepaspectratio]{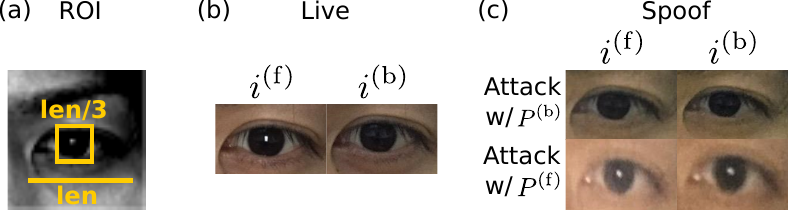}}
\caption{Specular reflections from iris regions. (a) Region of interest (ROI). A square ROI whose edge is one-third of the horizontal eye length (``len'' in the panel) is used to crop the iris region. (b) Live iris pictures taken with ($i^{(\mathrm f)}$) or without ($i^{(\mathrm b)}$) a flash. (c) Spoof iris pictures taken with ($i^{(\mathrm f)}$) or without ($i^{(\mathrm b)}$) a flash. Top row: a print attack with a picture taken without a flash ($P^{(\mathrm b)}$). Bottom row: a print attack with a picture taken with a flash ($P^{(\mathrm f)}$).}
\label{fig:Iris_explain}
\end{figure}
Unlike a printed photo or image shown on a display, the human iris shows specular reflection (due to its curved, glass beads-like structure) when light is flashed in front of it. Thus, if $P^{(\mathrm f)}$ is from a live face, a white spot reflecting the flash appears, whereas in the case of a live $P^{(\mathrm b)}$, a white spot does not appear (Fig.\ref{fig:Iris_explain}b). On the other hand, if $P^{(\mathrm f)}$ and $P^{(\mathrm b)}$ are from a spoof face, a white spot appears in neither of them, but if a flashed face is used as the spoof face, it appears in both of them (Fig.\ref{fig:Iris_explain}c). To utilize this difference as $\mathcal{D}_{\mathrm{spec}}$, the iris regions are extracted from the cropped face according to $Face_{\mathrm{loc}}$. The iris regions are defined as two square boxes centered on each eye, having an edge length that is one-third of the horizontal length of the eye (Fig.\ref{fig:Iris_explain}a). A Gaussian kernel with a two-pixel standard deviation is applied to each of the regions and then they are resized to $40 \times 40$ pixels. Hereafter, the extracted and resized iris regions from both eyes are denoted as $i^{(\mathrm f)} \in\mathbb{R}^{40 \times 40 \times 2}$ and $i^{(\mathrm b)} \in\mathbb{R}^{40 \times 40 \times 2}$. Pixel intensities at the vertical location $h$, horizontal location $w$, and eye position $s$ are denoted as $i^{(\mathrm f)}_{\mathrm{h,w,s}}$ and $i^{(\mathrm b)}_{\mathrm{h,w,s}}$. An intermediate descriptor $\mathcal{S}$ is calculated by pixel-wise subtraction of $i^{(\mathrm b)}_{\mathrm{h,w,s}}$ from $i^{(\mathrm f)}_{\mathrm{h,w,s}}$, which is then normalized by the sum of the luminance magnitudes, $i^{(\mathrm f)}_{\mathrm{h,w,s}} + i^{(\mathrm b)}_{\mathrm{h,w,s}}$, as follows:

\begin{equation}\label{normscore}
 \mathcal{S}_{\mathrm{h,w,s}} =
 \begin{cases}
   0 & \text{if $i^{(\mathrm f)}_{\mathrm{h,w,s}} = i^{(\mathrm b)}_{\mathrm{h,w,s}} = 0$} \\
   \displaystyle\frac{i^{(\mathrm f)}_{\mathrm{h,w,s}} - i^{(\mathrm b)}_{\mathrm{h,w,s}}}{i^{(\mathrm f)}_{\mathrm{h,w,s}} + i^{(\mathrm b)}_{\mathrm{h,w,s}}} & \text{otherwise.}
 \end{cases}
\end{equation}

\noindent Because $i^{(\mathrm f)}_{\mathrm{h,w,s}}$ and $i^{(\mathrm b)}_{\mathrm{h,w,s}}$ are greater than or equal to zero, $\mathcal{S}_{\mathrm{h,w,s}} \in [-1, 1]$.\\ 
\indent One potential weakness of $\mathcal{D}_{\mathrm{spec}}$ is its sensitivity to change in the position of the reflected-light spot. Depending on the relative position of the subject's face and direction of the flash, the position of the white-reflection spot inside the iris region changes. Although Gaussian filtering increases positional invariance, the variance of the spot position is much larger than the Gaussian-kernel width. Thus, to further increase positional invariance, the elements of the vectorized descriptor that are originated from each eye are sorted in ascending order to obtain $\mathcal{D}_{\mathrm{spec}}\in\mathbb{R}^{3200}$ as follows:
\begin{equation}\label{Dspeculum}
\begin{split}
\mathcal{S} &= vectorize(\mathcal{S})\\
\mathcal{D}_{\mathrm{spec}} &= sort(\mathcal{S})
\end{split}
\end{equation}

\subsection{$\mathcal{D}_{\mathrm{diff}}$: diffuse-reflection-based descriptor}\label{subsec:Ddiff}
Although it has been confirmed that $\mathcal{D}_{\mathrm{spec}}$ by itself can detect spoofing attacks, it has several pitfalls. Firstly, if a live subject is wearing glasses, the lens surface reflects the flash. The false-negative rate is increased when glasses-originated specular light contaminates the iris-originated specular light. Secondly, if a photo printed on a glossy paper is bent and used for an attack, there is a slight chance that the flash will reflect at the iris region of the printed photo, leading to increased false-negative rate. To compensate for this risk, we propose another liveness descriptor based on facial diffuse reflection, called the \textit{Diffusion Descriptor} or $\mathcal{D}_{\mathrm{diff}}$. $\mathcal{D}_{\mathrm{diff}}$ represents the surface structure of a face: live faces have the 3D structures, whereas spoof faces have 2D flat surfaces. An intermediate descriptor $\mathcal{S}$ is calculated from the pixel intensities in the face region (in a similar manner to equation \ref{normscore}) as follows:
\begin{equation}\label{Ddiffusion}
 \mathcal{S}_{\mathrm{h,w}} =
 \begin{cases}
   0 & \text{if $I^{(\mathrm f)}_{\mathrm{h,w}} = I^{(\mathrm b)}_{\mathrm{h,w}} = 0$} \\
   \displaystyle\frac{I^{(\mathrm f)}_{\mathrm{h,w}} - I^{(\mathrm b)}_{\mathrm{h,w}}}{I^{(\mathrm f)}_{\mathrm{h,w}} + I^{(\mathrm b)}_{\mathrm{h,w}}} & \text{otherwise}
 \end{cases}
\end{equation}
\noindent where $I^{(\mathrm f)} \in\mathbb{R}^{100 \times 100}$ and $I^{(\mathrm b)} \in\mathbb{R}^{100 \times 100}$ are face regions in photos $P^{(\mathrm f)}$ and $P^{(\mathrm b)}$, cropped with rectangles circumscribing all $Face_{\mathrm{loc}}$, filtered with a Gaussian kernel with a five-pixel standard deviation, and resized to $100 \times 100$ pixels. Here, $\mathcal{S}_{\mathrm{h,w}} \in [-1, 1]$ because $I^{(\mathrm f)}_{\mathrm{h,w}}$ and $I^{(\mathrm b)}_{\mathrm{h,w}}$ are greater than or equal to zero. Unlike the case of $\mathcal{D}_{\mathrm{spec}}$, the intermediate descriptor $\mathcal{S}$ is vectorized without sorting to preserve the spatial integrity of the face region:\\
\begin{equation}\label{StoDdiff}
\mathcal{D}_{\mathrm{diff}} = vectorize(\mathcal{S}).
\end{equation}
\indent In light of the Lambertian model, we can understand why does $\mathcal{D}_{\mathrm{diff}}$ represent the 3D structure of a face. Moreover, the Lambertian model explains an additional advantage of $\mathcal{D}_{\mathrm{diff}}$: color invariance (also see \cite{MDM}). This can be seen as follows: assuming that the entire face is a Lambertian surface (i.e., a uniform diffuser), the surface-luminance intensity depends on the radiant intensity per unit area in the direction of observation. Thus, the pixel intensity $I_{\mathrm{h,w}}$ at the vertical position $h$ and horizontal position $w$ can be described as:

\begin{equation}\label{Lambert}
 I_{\mathrm{h,w}} = LK_{\mathrm{h,w}}\cos\theta_{\mathrm{h,w}}
\end{equation}

\noindent where $L$, $K_{\mathrm{h,w}}$, and $\theta_{\mathrm{h,w}}$  denote the light-source intensity, surface reflectance coefficient, and angle of incidence, respectively. As equation \ref{Lambert} indicates, the luminance intensity $I_{\mathrm{h,w}}$ depends on the 3D structure of the facial surface that determines $\theta_{\mathrm{h,w}}$. Additionally, $I_{\mathrm{h,w}}$ depends on the surface reflectance $K_{\mathrm{h,w}}$. This means that differences in color of the surface (e.g., light skin vs. dark skin) affect the observed luminance intensity even under the same light intensity, $L$. The design of equation \ref{Ddiffusion} solves color-dependency problem by canceling out the surface reflectance $K_{\mathrm{h,w}}$. Under the assumption of a Lambertian surface, the terms $I^{(\mathrm f)}_{\mathrm{h,w}}$ and $I^{(\mathrm b)}_{\mathrm{h,w}}$ are expressed as:	

\begin{equation}\label{flashBG}
\begin{split}
 I^{(\mathrm f)}_{\mathrm{h,w}} &= L^{(\mathrm f)}K_{\mathrm{h,w}}\cos\theta_{\mathrm{h,w}} + L^{(\mathrm f)}K_{\mathrm{h,w}} \\
 I^{(\mathrm b)}_{\mathrm{h,w}} &= L^{(\mathrm b)}K_{\mathrm{h,w}}
\end{split}
\end{equation}

\noindent where $L^{(\mathrm f)}$ and $L^{(\mathrm b)}$ are the intensities of the flash light and background light (ambient light), respectively. Since ambient light coming from all directions is integrated, the background-light term does not depend on the incident angle of the light. Substituting $I^{(\mathrm f)}_{\mathrm{h,w}}$ and $I^{(\mathrm b)}_{\mathrm{h,w}}$ into equation \ref{Ddiffusion} yields the intermediate descriptor $\mathcal{S}$:

\begin{equation}\label{Ddiffusion2}
 \mathcal{S}_{\mathrm{h, w}} =
 \begin{cases}
   0 & \text{if $I^{(\mathrm f)}_{\mathrm{h,w}} = I^{(\mathrm b)}_{\mathrm{h,w}} = 0$} \\
   \displaystyle\frac{L^{(\mathrm f)}\cos\theta_{\mathrm{h,w}}}{L^{(\mathrm f)}\cos\theta_{\mathrm{h,w}} + 2L^{(\mathrm b)}} & \text{otherwise}.
 \end{cases}
\end{equation}

\noindent Equation \ref{Ddiffusion2} depends on $\theta$ and represents the 3D structure of the facial surface. Yet equation \ref{Ddiffusion2} is independent of the surface reflectance $K$, thereby avoiding the skin-color problem. Thus, although Lambertian reflections from the facial surface can be modeled as a function of the surface reflectance and surface 3D structure, equation \ref{Ddiffusion} cancels $K$ in order to confer color invariance as an additional advantage to $\mathcal{D}_{\mathrm{diff}}$. 

\begin{figure}[tb]
\centerline{\includegraphics[keepaspectratio]{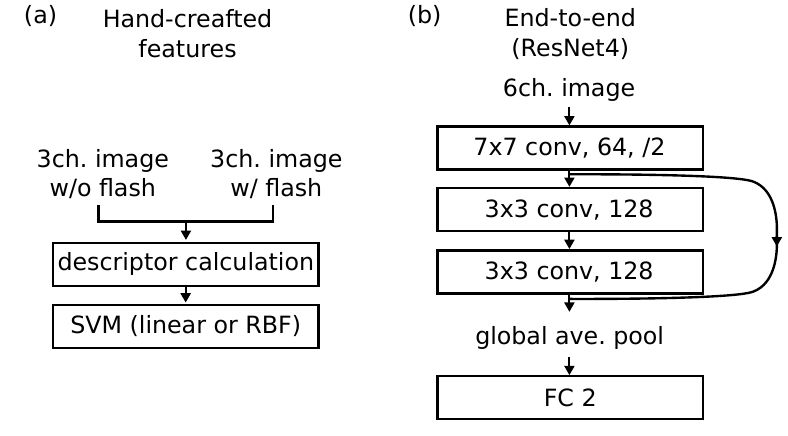}}
\caption{Classifiers. (a) SVM classifier applied to the calculated descriptors, $\mathcal{D}_{\mathrm{spec}}$, $\mathcal{D}_{\mathrm{diff}}$, and $SpecDiff$. (b) ResNet4, an end-to-end classifier taking $I'^{(\mathrm f)}$ and $I'^{(\mathrm b)}$ as a six-channel tensor.}
\label{fig:classifier}
\end{figure}

\subsection{$SpecDiff$ descriptor}
\noindent The two descriptors are concatenated into the $SpecDiff$ descriptor:
\begin{equation}\label{cat_SpecDiff}
SpecDiff = concatenate(\mathcal{D}_{\mathrm{spec}}, \mathcal{D}_{\mathrm{diff}})
\end{equation}
\noindent The $SpecDiff$ descriptor attains higher classification accuracy compared with either $\mathcal{D}_{\mathrm{spec}}$ or $\mathcal{D}_{\mathrm{diff}}$ alone. As both $\mathcal{D}_{\mathrm{spec}}$ and $\mathcal{D}_{\mathrm{diff}}$ are normalized in the range $[-1, 1]$, $SpecDiff$ also has a bounded descriptor magnitude that helps model training and classification. The experiments described below, however, test not only $SpecDiff$; the ablation studies test $\mathcal{D}_{\mathrm{spec}}$ and $\mathcal{D}_{\mathrm{diff}}$ by themselves.
\section{Experiments}
\subsection{Databases}
As of late-2019, there is no publicly available facial-image database of images taken with and without a flash. Therefore, we collected 1176 and 1660 photos of live and spoof faces of 20 subjects, respectively. The images are taken under two lighting conditions (bright office area or dark corridor area), two facial-accessory conditions (glasses or nothing), and three facial expressions (smile, mouth opened, or no expression). Each condition has 4 or 5 repeats with varying backgrounds. Thus, for the live faces, one subject has approximately 60 repeats. Spoof faces include flat-papers, bent-papers, and displays. For the flat-paper conditions, one subject has the same number of repeats as the live faces. For the bent-paper conditions, one subject has approximately 12 repeats, including 4 bend directions (horizontal, vertical, and two diagonal directions) and at least one lighting condition (bright office area), two facial-accessory conditions (glasses or nothing), and two facial expressions (two of the three expressions as above). For the display conditions, one subject has approximately 10 repeats, including at least one lighting condition (bright office area), two facial-accessory conditions (glasses or nothing), three facial expressions (smile, mouth opened, or no expression). Thus, for the spoof faces, one subject has at least 82 repeats in total. \\
\indent To test the generalizability of the proposed method, a cross-database validation is conducted using presentation attacks contained in the three public databases, NUAA, Replay-Attack, and SiW (test subset), which consist of 15 IDs / 7509 pictures, 50 IDs / 1000 videos, and 75 IDs / 1462 videos, respectively, simulating photo and display attacks. For each photo and video, we take images with and without a flash to create test databases. These public databases are used exclusively for testing models. For both in-house and public databases, the device used for the data collection is an iPhone7 (A1779), and the display-attack devices are an iPad Pro (A1584) and ASUS ZenFone Go (ZB551KL).
\subsection{Models}
The classification performances of the proposed descriptors are evaluated using SVM, either with a linear or radial basis function (RBF) kernel (Fig.\ref{fig:classifier}a). Libsvm package is used for the SVM training \cite{libsvm}. $\mathcal{D}_{\mathrm{spec}}$, $\mathcal{D}_{\mathrm{diff}}$, and $SpecDiff$ are compared with four previously reported flash-based descriptors: SD\textunderscore FIC ($\mathbb{R}$, \cite{FlashLivenessChan}), LBP\textunderscore FI ($\mathbb{R}^{10000}$, \cite{FlashLivenessChan}), RelativeRef ($\mathbb{R}^{10000}$, \cite{FlashLivenessTang}), and Implicit3D ($\mathbb{R}^{10000}$, \cite{MDM}).\\
\indent ResNet4 (Fig.\ref{fig:classifier}b) and ResNet18 \cite{ResNetV2} are constructed as end-to-end deep neural networks. They take as input a six-channel image that is constructed with two three-channel images $I'^{(\mathrm f)}$ and $I'^{(\mathrm b)}$ concatenated along the channel axis. $I'^{(\mathrm f)}$ and $I'^{(\mathrm b)}$ are facial images with and without a flash, undergone the same preprocessing steps as the proposed algorithm, except \textit{GrayScale} and \textit{GaussFilter} functions. The resulting input image size is $[244 \times 244 \times 6]$. the last two layers, global average pooling and FC layer of ResNet4 and ResNet18, classify a 128-channel tensor into one of the two alternative classes. We do not consider neural networks that are deeper or more complex than ResNet18, because ResNet18 shows significantly worse results compared with that of ResNet4 (See Results section).
\subsection{Evaluation metrics}
Following ISO/IEC 30107-3 metrics, we calculated attack presentation classification error rate (APCER), bona fide presentation classification error rate (BPCER), and average classification error rate (ACER). \\
\indent One problem in evaluation is that we cannot access to the live subjects appear in the public databases. Thus, we can only evaluate APCER using spoof subjects either printed on papers or projected on screens. In order to obtain metrics equivalent to BPCER and ACER, we isolated a part of the live faces of in-house database from the training dataset and used them as a substitute for live faces of the public databases. Although the live faces from the in-house database are not exactly same as that of public databases, simulated BPCER and ACER serve as conservative measures of the true BPCER and ACER, because our in-house database contains a variety of conditions such as lighting, facial accessory, and expression. This heterogeneity leads to a broad, heavy-tailed distribution of live face scores from the in-house database.  Hereafter we mention these simulated metrics as simulated BPCER (sBPCER) and simulated ACER (sACER).
\begin{figure}[tb]
\centerline{\includegraphics[keepaspectratio,width=8cm]{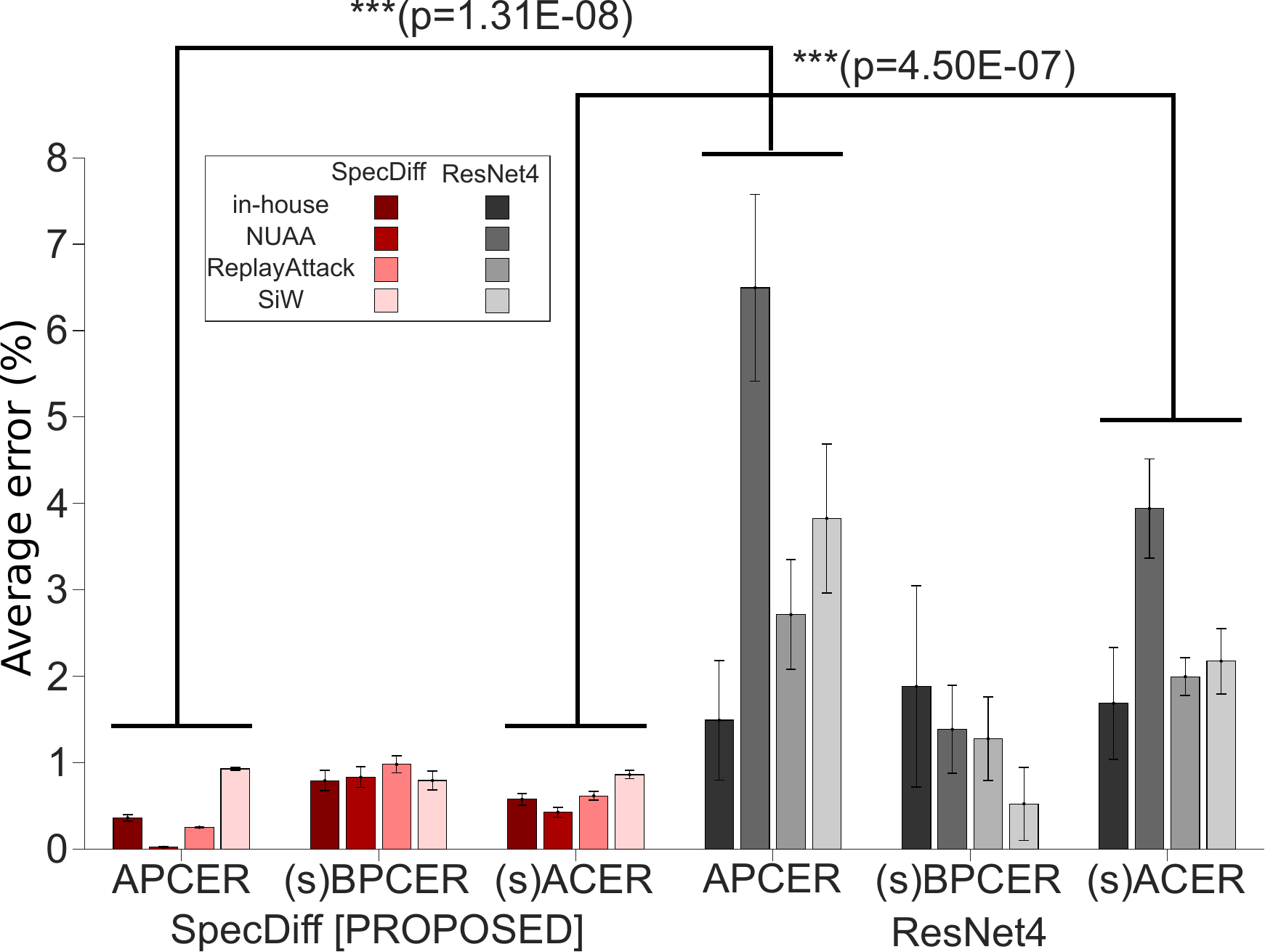}}
\caption{Two-way ANOVA comparing $SpecDiff$ and ResNet4. (s)BPCER and (s)ACER indicate that sBPCER and sACER are used for evaluating public databases, while genuine BPCER and ACER are used for evaluating the in-house database. Resulting p-values show statistical significance in APCER and (s)ACER.}
\label{fig:ANOVA}
\end{figure}
\subsection{Speed test on actual mobile devices}
To compare the execution speeds of our proposed algorithm and ResNet4 under real-world scenarios, we built a custom-made iOS application for PAD on Xcode 10.2.1 / MacBook Pro, written in Swift, C, and C$\sharp$. For Gaussian filtration, an OpenCV \cite{opencv} built-in function is used. The app is then installed on an iPhone7 (A1779), iPhone XR (A2106), and iPad Pro (A1876) for the speed evaluation. Execution speed is measured during the preprocessing step and the descriptor calculation/classification step.
\begin{figure*}[tb]
\centerline{\includegraphics[keepaspectratio,width=17.5cm]{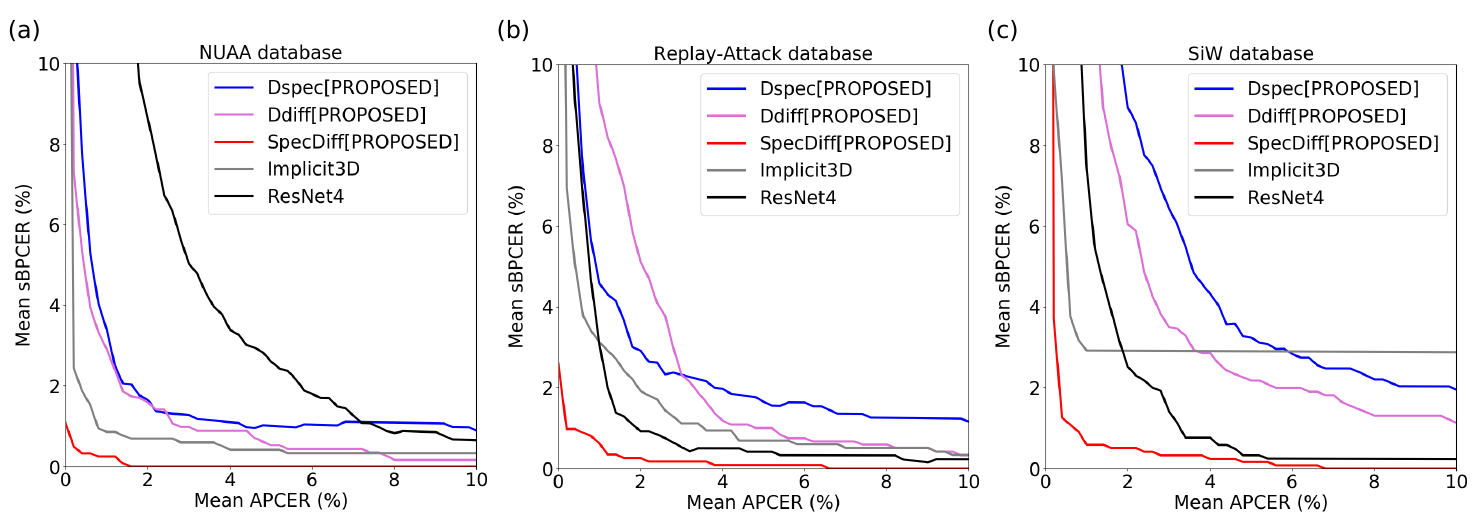}}
\caption{Mean ROC curves of the cross-validation experiments. Note that only [0\%, 10\%] range is shown out of [0\%, 100\%]. (a) 10-fold cross-validation on the NUAA database. (b) 10-fold cross-validation on the Replay-Attack database. (c) 10-fold cross-validation on the SiW database.}
\label{fig:ROC}
\end{figure*}
\begin{figure}[tb]
\centerline{\includegraphics[keepaspectratio]{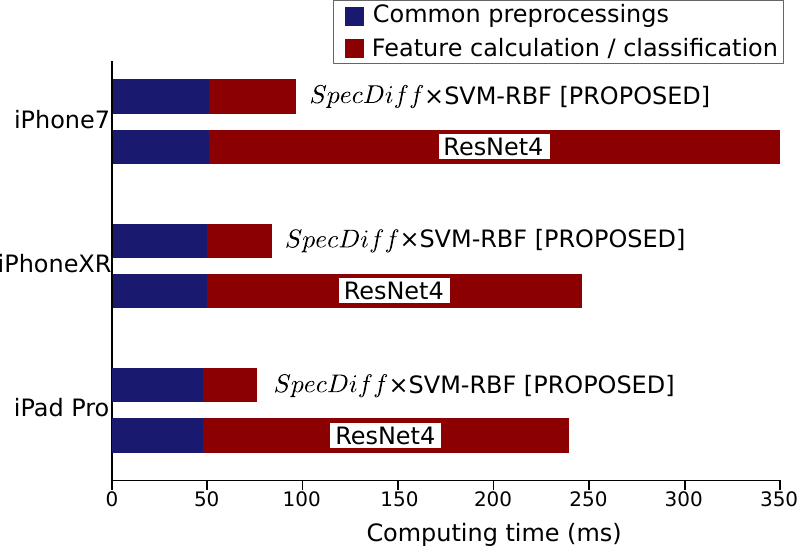}}
\caption{Summary of execution speeds. The proposed $SpecDiff$ descriptor classified with the SVM-RBF kernel is compared with ResNet4. Execution speeds on iPhone7, iPhone XR, and iPad Pro are measured.}
\label{fig:speed}
\end{figure}
\section{Results}
\subsection{Classification performance}
A leave-one-ID-out cross-validation is conducted on the in-house database. At each validation cycle, all data from one subject out of 20 IDs are separated as a test dataset. The remaining data from 19 IDs are used to train the models and to determine thresholds. The results are summarized in Table\ref{tab:LeaveOneOutXval}, with the best model highlighted in bold. The average $\mathcal{D}_{\mathrm{spec}}$ and $\mathcal{D}_{\mathrm{diff}}$ calculated using the in-house database are shown in Fig.\ref{fig:descriptor}a and b, respectively. \\
\indent To evaluate classification performances on the public databases, we conduct 10-fold cross-validation. At each validation, model training and threshold determination are conducted with randomly selected $90\%$ of the in-house database, while live faces of the remaining $10\%$ together with spoof faces from the public databases are used to compute sBPCER and APCER. The results of the NUAA, Replay-Attack, and SiW databases are summarized in Table \ref{tab:CrossDataset_NUAA}, \ref{tab:CrossDataset_ReplayAttack}, and \ref{tab:CrossDataset_SiW}. Fig.\ref{fig:ROC} shows the mean ROC curves of the top five models.\\
\indent Among all the descriptors-classifiers combination, the proposed $SpecDiff$ descriptor with RBF kernel-SVM achieves the highest classification accuracy on average, on both the in-house database and on the three public databases. Moreover, its accuracy is statistically significantly better than ResNet4, the end-to-end deep-neural-network classifier. We conduct two-way ANOVA with the two factors, model and database, to test if APCER, BPCER, and ACER are significantly different. Model factor contains two members, $SpecDiff$ and ResNet4, and database factor contains four members, in-house, NUAA, ReplayAttack, and SiW. The resulting p-values of APCER and ACER are less than $0.001$, indicating that $SpecDiff$ descriptor with SVM achieves significantly better accuracy than ResNet4 (Fig.\ref{fig:ANOVA}). Following Tukey-Kramer multi-comparison test show that $SpecDiff$ achieves significantly smaller APCER and ACER than ResNet4 on each database ($p<0.001$).\\
\indent We observe statistically significant increase in APCER, ACER and sACER using ResNet18, compared with the shallower network, ResNet4 ($p<0.05$). This is potentially due to the lack of a large database: generally speaking, a deeper neural network can benefit from its higher complexity and nonlinearity to achieve a better performance, but with our small in-house database, a simpler classifier with properly designed feature performs better while a complex classifier with a large number of parameters suffers from overfitting.\\
\indent To test vulnerability to the two presentation attack instruments, photo and display, the $SpecDiff$ descriptor with the RBF-kernel SVM is tested separately against photo and replay attacks by using the in-house, Replay-Attack, and SiW database. Leave-one-ID-out-cross-validation is performed on the in-house database. For the two public databases, SVM is trained using an entire in-house database and tested with the public database. As summarized in table \ref{tab:photoVSdisplay}, APCER of the proposed model on display attacks of public databases is comparable to its performance on photo attacks, although on the in-house database video APCER is slightly worse (one-way ANOVA, $p<0.05$). \\
\indent To test vulnerability to the lighting conditions, we conducted Leave-one-ID-out-cross-validation on the in-house database, tested separately on the data taken in a bright office area, and in a dark corridor area. The result summarized in table \ref{tab:brightVSdark} shows that variation in lighting conditions do not significantly alter the classification performances (one-way ANOVA, $p>0.05$). 

\subsection{Execution speed on mobile devices}
The results of the speed evaluation of the proposed algorithm and deep neural network classifier on the iPhone7, iPhone XR and iPad Pro are summarized in Fig.\ref{fig:speed}. On all devices, the proposed algorithm is approximately six-times faster in terms of the descriptor calculation/classification time than ResNet4.

\begin{table}[htbp]
  \centering
  \footnotesize
  \caption{In-house database cross-validation errors ($\%$).}
    \begin{tabular}{cccccc}
    Descriptor & \multicolumn{2}{c}{Classifier} & APCER & BPCER & ACER\\
    \midrule
    \multicolumn{1}{c}{\multirow{2}[1]{*}{SD\_FIC \cite{FlashLivenessChan}}} & \multirow{2}[1]{*}{SVM} & linear & 26.15 & 47.36 & 36.76 \\
          &       & RBF & 24.85 & 48.43 & 36.64 \\
    \multicolumn{1}{c}{\multirow{2}[0]{*}{LBP\_FI \cite{FlashLivenessChan}}} & \multirow{2}[0]{*}{SVM} & linear & 4.09 & 12.81 & 8.45 \\
          &       & RBF   & 1.73 & 10.53 & 6.13 \\
    Relative- & \multirow{2}[0]{*}{SVM} & linear & 2.11 & 80.31 & 41.21 \\
    Ref \cite{FlashLivenessTang}      &       & RBF   & 1.72 & 28.02 & 14.87 \\
    \multicolumn{1}{c}{\multirow{2}[1]{*}{Implicit3D \cite{MDM}}} & \multirow{2}[1]{*}{SVM} & linear & 21.48 & 24.36 & 22.92 \\
          &       & RBF   & 1.10 & 2.17 & 1.63  \\
    \midrule
    $\mathcal{D}_{\mathrm{spec}}$  & \multirow{2}[1]{*}{SVM} & linear & 1.15 & 2.94 & 2.04 \\
    {[PROPOSED]}      &       & RBF   & 0.74 & 2.44 & 1.59 \\
    $\mathcal{D}_{\mathrm{diff}}$  & \multirow{2}[0]{*}{SVM} & linear & 1.67 & 2.89 & 2.28 \\
    {[PROPOSED]}      &       & RBF   & 0.67 & 1.81 & 1.24 \\
    \multicolumn{1}{c}{\multirow{2}[1]{*}{\shortstack[1]{$SpecDiff$\\ {[PROPOSED]}}}} & \multirow{2}[0]{*}{SVM} & linear & 0.62 & 1.30 & 0.96 \\
          &       & RBF   & \textbf{0.36} & \textbf{0.79} & \textbf{0.58} \\
    \midrule
    \multicolumn{3}{c}{ResNet4} & 1.49 & 1.88 & 1.69 \\
    \multicolumn{3}{c}{ResNet18} & 9.72 & 2.25 & 5.98 \\
    \bottomrule
    \end{tabular}%
    \label{tab:LeaveOneOutXval}%
\end{table}%

\begin{table}[htbp]
  \centering
  \footnotesize
  \caption{NUAA cross-database validation errors ($\%$).}
    \begin{tabular}{cccccc}
    Descriptor & \multicolumn{2}{c}{Classifier} & APCER & sBPCER & sACER \\
    \midrule
    \multicolumn{1}{c}{\multirow{2}[1]{*}{SD\_FIC \cite{FlashLivenessChan}}} & \multirow{2}[1]{*}{SVM} & linear & 18.58 & 56.90 & 37.34\\
          &       & RBF   & 27.21 & 50.66 & 38.94  \\
    \multicolumn{1}{c}{\multirow{2}[0]{*}{LBP\_FI \cite{FlashLivenessChan}}} & \multirow{2}[0]{*}{SVM} & linear & 40.14 & 7.73 & 23.94  \\
          &       & RBF   & 21.55 & 16.49 & 19.02   \\
    Relative- & \multirow{2}[0]{*}{SVM} & linear & 23.12 & 2.85 & 12.98  \\
    Ref \cite{FlashLivenessTang}      &       & RBF   & 10.30 & 3.59 & 6.95 \\
    \multicolumn{1}{c}{\multirow{2}[1]{*}{Implicit3D \cite{MDM}}} & \multirow{2}[1]{*}{SVM} & linear & 0.62 & 3.24 & 1.93 \\
          &       & RBF   & 0.42 & 2.06 & 1.24\\
    \midrule
    $\mathcal{D}_{\mathrm{spec}}$  & \multirow{2}[1]{*}{SVM} & linear & 1.00 & 2.26 &  1.63\\
    {[PROPOSED]}      &       & RBF   & 1.61 & 2.03 & 1.82 \\
    $\mathcal{D}_{\mathrm{diff}}$ & \multirow{2}[0]{*}{SVM} & linear & 0.84 & 2.05 & 1.45\\
    {[PROPOSED]}      &       & RBF   & 1.12 & 3.07 & 2.10\\
    \multicolumn{1}{c}{\multirow{2}[1]{*}{\shortstack[1]{$SpecDiff$\\ {[PROPOSED]}}}} & \multirow{2}[0]{*}{SVM} & linear & 0.052 & 0.91 & 0.48 \\
          &       & RBF   & \textbf{0.021} & \textbf{0.83} & \textbf{0.43}\\
    \midrule
    \multicolumn{3}{c}{ResNet4} & 6.50 & 1.39 & 3.94 \\
    \multicolumn{3}{c}{ResNet18} & 55.61 & 9.07 & 32.34 \\
    \bottomrule
    \end{tabular}%
    \label{tab:CrossDataset_NUAA}%
\end{table}%

\begin{table}[htbp]
  \centering
  \footnotesize
  \caption{Replay Attack cross-database validation errors ($\%$).}
    \begin{tabular}{cccccc}
    Descriptor & \multicolumn{2}{c}{Classifier} & APCER & sBPCER & sACER\\
    \midrule
    \multicolumn{1}{c}{\multirow{2}[1]{*}{SD\_FIC \cite{FlashLivenessChan}}} & \multirow{2}[1]{*}{SVM} & linear & 40.62 & 58.15 & 49.38 \\
          &       & RBF   & 33.22 & 49.98 & 41.60 \\
    \multicolumn{1}{c}{\multirow{2}[0]{*}{LBP\_FI \cite{FlashLivenessChan}}} & \multirow{2}[0]{*}{SVM} & linear & 35.79 & 6.02 & 20.90 \\
          &       & RBF   & 2.11 & 22.07 & 12.09 \\
    Relative- & \multirow{2}[0]{*}{SVM} & linear & 27.74 & 2.12 & 14.93 \\
    Ref\cite{FlashLivenessTang}      &       & RBF   & 6.57 & 3.98 & 5.27 \\
    \multicolumn{1}{c}{\multirow{2}[1]{*}{Implicit3D \cite{MDM}}} & \multirow{2}[1]{*}{SVM} & linear & 6.86 & 2.17 & 4.51 \\
          &       & RBF   & 2.24 & 2.06 & 2.15 \\
    \midrule
    $\mathcal{D}_{\mathrm{spec}}$ & \multirow{2}[1]{*}{SVM} & linear & 5.24 & 3.62 & 4.43 \\
    {[PROPOSED]}      &       & RBF   &  1.95 & 2.35 & 2.15 \\
    $\mathcal{D}_{\mathrm{diff}}$  & \multirow{2}[0]{*}{SVM} & linear & 9.56 & 2.61 & 6.08 \\
    {[PROPOSED]}      &       & RBF   & 3.51 & 1.28 & 2.39  \\
    \multicolumn{1}{c}{\multirow{2}[1]{*}{\shortstack[1]{$SpecDiff$\\ {[PROPOSED]}}}} & \multirow{2}[0]{*}{SVM} & linear & 2.73 & \textbf{0.95} & 1.84 \\
          &       & RBF   & \textbf{0.25} & 0.98 & \textbf{0.62} \\
    \midrule
    \multicolumn{3}{c}{ResNet4} & 2.71 & 1.28 & 1.99 \\
    \multicolumn{3}{c}{ResNet18} & 35.32 & 0.97 & 18.14\\
    \bottomrule
    \end{tabular}%
    \label{tab:CrossDataset_ReplayAttack}%
\end{table}%

\begin{table}[htbp]
  \centering
  \footnotesize
  \caption{SiW cross-database validation errors ($\%$).}
    \begin{tabular}{cccccc}
    Descriptor & \multicolumn{2}{c}{Classifier} & APCER & sBPCER & sACER\\
    \midrule
    \multicolumn{1}{c}{\multirow{2}[1]{*}{SD\_FIC \cite{FlashLivenessChan}}} & \multirow{2}[1]{*}{SVM} & linear & 51.53 & 58.10 & 54.81 \\
          &       & RBF   & 44.87 & 47.76 & 46.32 \\
    \multicolumn{1}{c}{\multirow{2}[0]{*}{LBP\_FI \cite{FlashLivenessChan}}} & \multirow{2}[0]{*}{SVM} & linear & 14.98 & 5.43 & 10.21 \\
          &       & RBF   & 3.36 & 21.95 & 12.66 \\
    Relative- & \multirow{2}[0]{*}{SVM} & linear & 57.6 & 5.08 & 31.36 \\
    Ref\cite{FlashLivenessTang}      &       & RBF   & 25.68 & 3.98 & 14.83 \\
    \multicolumn{1}{c}{\multirow{2}[1]{*}{Implicit3D \cite{MDM}}} & \multirow{2}[1]{*}{SVM} & linear & 9.40 & 3.01 & 6.21 \\
          &       & RBF   & 36.16 & 0.62 & 18.39 \\
    \midrule
    $\mathcal{D}_{\mathrm{spec}}$ & \multirow{2}[1]{*}{SVM} & linear & 16.66 & 2.76 & 9.70 \\
    {[PROPOSED]}      &       & RBF   &  5.79 & 2.66 & 4.23 \\
    $\mathcal{D}_{\mathrm{diff}}$  & \multirow{2}[0]{*}{SVM} & linear & 5.79 & 1.84 & 3.82 \\
    {[PROPOSED]}      &       & RBF   & 3.61 & 2.71 & 3.16 \\
    \multicolumn{1}{c}{\multirow{2}[1]{*}{\shortstack[1]{$SpecDiff$\\ {[PROPOSED]}}}} & \multirow{2}[0]{*}{SVM} & linear & 2.86 & 0.91 & 1.88 \\
          &       & RBF   & \textbf{0.93} & 0.79 & \textbf{0.86} \\
    \midrule
    \multicolumn{3}{c}{ResNet4} & 3.83 & 0.52 & 2.17 \\
    \multicolumn{3}{c}{ResNet18} & 21.74 & \textbf{0.18} & 10.96 \\
    \bottomrule
    \end{tabular}%
    \label{tab:CrossDataset_SiW}%
\end{table}%

\begin{table}[htbp]
  \centering
  \footnotesize
  \caption{Evaluation on spoofing subcategories by using $SpecDiff$ descriptor with SVM-RBF kernel classifier.}
    \begin{tabular}{ccc}
    Database & photo APCER & video APCER \\
    \midrule
    in-house mean & 0.19  & 2.72 \\
    Replay & 0.34  & 0.00 \\
    SiW   & 0.00   & 0.00 \\
    \bottomrule
    \end{tabular}%
  \label{tab:photoVSdisplay}%
\end{table}%

\begin{table}[htbp]
  \centering
  \footnotesize
  \caption{Evaluation on different lighting conditions by using the in-house database and $SpecDiff$ descriptor with SVM-RBF kernel classifier.}
    \begin{tabular}{cccc}
    Lighting & APCER & BPCER & ACER \\
    \midrule
    Bright & 0.27  & 0.94  & 0.61 \\
    Dark  & 0.0087 & 0.28  & 0.15 \\
    \bottomrule
    \end{tabular}%
  \label{tab:brightVSdark}%
\end{table}%

\section{Conclusion}
By using specular and diffusion reflection from a subject's face, the proposed algorithm based on the $SpecDiff$ descriptor achieved the best PAD accuracy among other flash-based algorithms at execution speed approximately six-times faster than that of a deep neural network. The algorithm requires only one visible-light camera and a flash light. A small database containing $\approx 1K$ image pairs per class with binary labels is sufficient to train a classifier using the $SpecDiff$ descriptor, enabling the easy and wide application of the PAD algorithm. Experiments conducted on the algorithm operating on actual devices confirms that it has a practical level of performance on mobile devices without the need for computationally expensive processing units.

\section*{Acknowledgements}
The authors thank the anonymous reviewers for their careful reading to improve the manuscript. We would also like to thank Koichi Takahashi, Kazuo Sato, Yoshitoki Ideta, Taiki Miyagawa, and Yuka Fujii for the insightful discussions and supports of the project.

{\small
\bibliographystyle{ieee}
\bibliography{Liveness_BibTeX}
}

\end{document}